\documentclass{article}

\usepackage[preprint,nonatbib]{neurips_2020}

\usepackage[utf8]{inputenc} 
\usepackage[T1]{fontenc}    
\usepackage{hyperref}       
\usepackage{url}            
\usepackage{booktabs}       
\usepackage{amsfonts}       
\usepackage{nicefrac}       
\usepackage{microtype}      

\usepackage{graphicx}
\usepackage{multirow}
\usepackage{subcaption}
\usepackage{textcomp}
\usepackage{xcolor}

\usepackage{tikz}
\usetikzlibrary{shapes.geometric, arrows, positioning}

\tikzstyle{startstop} = [rectangle, rounded corners, 
minimum width=2cm,
minimum height=1cm,
text centered, 
text width=2cm,
draw=black]

\tikzstyle{process} = [rectangle, 
minimum width=2cm,
minimum height=1cm, 
text centered, 
text width=2cm,
draw=black]

\tikzstyle{arrow} = [thick,->,>=stealth]

\title{Anomaly Detection of Command Shell Sessions based on DistilBERT: Unsupervised and Supervised Approaches}

\author{
    Zefang Liu \\
    JPMorgan Chase \\
    \texttt{zefang.liu@jpmchase.com} \\
    \And
    John Buford \\
    JPMorgan Chase \\
    \texttt{john.buford@chase.com} \\
}

\begin{document}

\maketitle

\begin{abstract}
    Anomaly detection in command shell sessions is a critical aspect of computer security. Recent advances in deep learning and natural language processing, particularly transformer-based models, have shown great promise for addressing complex security challenges. In this paper, we implement a comprehensive approach to detect anomalies in Unix shell sessions using a pretrained DistilBERT model, leveraging both unsupervised and supervised learning techniques to identify anomalous activity while minimizing data labeling. The unsupervised method captures the underlying structure and syntax of Unix shell commands, enabling the detection of session deviations from normal behavior. Experiments on a large-scale enterprise dataset collected from production systems demonstrate the effectiveness of our approach in detecting anomalous behavior in Unix shell sessions. This work highlights the potential of leveraging recent advances in transformers to address important computer security challenges.
\end{abstract}

\section{Introduction}

The complexity of modern computer systems and networks has led to an increasing demand for efficient and reliable security solutions. Interactive command shells, especially Unix shells, which provide a powerful interface for system administration, development, and maintenance tasks, are an essential aspect of many computing environments. However, they can also be exploited by attackers to gain unauthorized access, escalate privileges, avoid defense detection, collect sensitive data, and manipulate systems. As a result, anomaly detection in command shells has become a crucial component of computer security.

Previous studies have utilized various techniques for anomaly detection in command shell sessions, ranging from simple rule-based methods to more complex machine learning algorithms. However, most of these approaches rely heavily on predefined features or labeled data from security experts for training supervised models. Assembling a large, well-labeled dataset can be time-consuming and labor-intensive, often resulting in a limited scope of detection capabilities due to the inherent biases in the labeling process.

Recent advances in deep learning and natural language processing (NLP) have enabled new opportunities for addressing complex security challenges. In particular, transformer-based models, such as BERT (Bidirectional Encoder Representations from Transformers) \cite{devlin2018bert} and GPT (Generative Pretrained Transformer) \cite{radford2018improving}, have achieved state-of-the-art performance across various NLP tasks. These models have the potential to enhance computer security by enabling more effective and adaptable anomaly detection systems that can learn from large-scale, diverse data sources.

In enterprise production environments, access to command shells is treated as a privileged activity because of the potential for misuse of system commands. Commands with the potential for misuse are well known. Specific commands may be a priori disabled. Attack techniques have been compiled, for example, in the MITRE ATT\&CK\textsuperscript{\textregistered} framework. Enterprises can implement rule-based detection using these resources. Consequently, the benefit of the anomaly detection model is to automatically identify command patterns that are outliers with respect to the overall set of sessions that would not be detected by the rule-based approach. Due to the volume, length, and complexity of shell sessions, manual detection of outliers is impractical. An automatic process is needed to assign anomaly scores to sessions, where sessions with high anomaly scores can be prioritized for further investigation.  In this paper, we apply a transformer-based model for anomaly detection in Unix shell sessions with a pretrained DistilBERT model. Our method employs both unsupervised and supervised learning techniques, aiming to deliver a robust and flexible solution for identifying anomalous activity while reducing the burden on manual labels from experts.  

\begin{figure}[!h]
  \centering
  \footnotesize
  \begin{tikzpicture}[node distance=3cm and 2cm]
    \node (start) [startstop] {Raw Keystroke Data};
    \node (pro1) [process, right of=start] {Unix Prompt Extraction};
    \node (pro2) [process, right of=pro1] {Unix Command Extraction};
    \node (pro3) [process, right of=pro2] {Unix Command Cleaning};
    
    \node (pro4) [process, below=0.5cm of start] {DistilBERT Pretraining};
    \node (pro5) [process, right of=pro4] {Unix Session Embedding};
    \node (pro6) [process, right of=pro5] {Unsupervised Anomaly Detection};
    \node (stop1) [startstop, right of=pro6] {Annotated Suspicious Sessions};
;
    \node (pro11) [process, below=0.5cm of pro5] {DistilBERT Fine-Tuning with SetFit};
    \node (pro12) [process, right of=pro11] {Supervised Session Classification};
    \node (stop2) [startstop, right of=pro12] {Predicted Suspicious Sessions};
    
    \draw [arrow] (start) -- (pro1);
    \draw [arrow] (pro1) -- (pro2);
    \draw [arrow] (pro2) -- (pro3);
    
    \draw [arrow] (pro3) |- ([shift={(0,-0.25cm)}]pro3.south) -| (pro4);
    \draw [arrow] (pro4) -- (pro5);
    \draw [arrow] (pro5) -- (pro6);
    \draw [arrow] (pro6) -- (stop1);

    \draw [arrow] (pro4) |- (pro11);
    \draw [arrow] (pro11) -- (pro12);
    \draw [arrow] (pro12) -- (stop2);
    \end{tikzpicture}
  \caption{Pipeline of the command shell session anomaly detection with both unsupervised and supervised methods.}
  \label{fig:pipeline}
\end{figure}
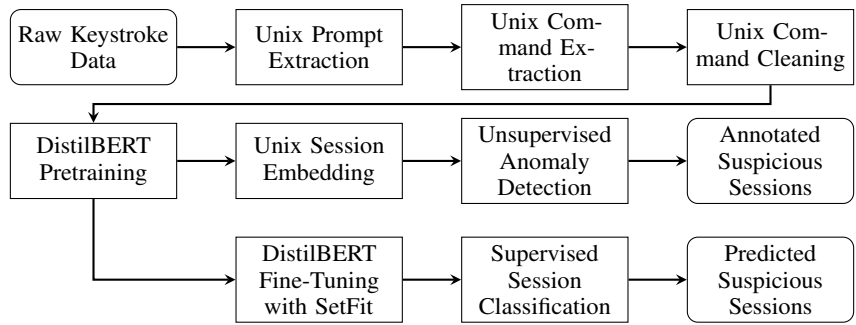

DistilBERT \cite{sanh2019distilbert}, a lighter and more efficient version of the BERT \cite{devlin2018bert}, has demonstrated exceptional performance across a wide range of NLP tasks. By pretraining a DistilBERT model on a large dataset of Unix shell sessions, we capture the underlying structure and syntax of Unix shell commands and allow the model to identify deviations of shell sessions from normal activity. The unsupervised method uses an ensemble model to calculate anomaly scores, detecting potential security threats without requiring labeled data. We further experimented with applying the unsupervised model to specific command subshells, such as HDFS, SQL, Spark, and Python, which are notable for having specific subshell command syntaxes. To further enhance the precision of our anomaly detection system, we implement a supervised approach by fine-tuning the pretrained DistilBERT model on a small set of labeled Unix shell sessions with suspicious keywords, which allows the model to learn from session labels and distinguish normal and anomalous activity more effectively. The overall pipeline is shown in the Figure \ref{fig:pipeline} for both unsupervised ans supervised methods.

The main contributions of this paper are as follows:
\begin{enumerate}
    \item We apply a comprehensive anomaly detection framework for Unix shell sessions based on the pretrained DistilBERT model and ensemble anomaly detectors, addressing an important problem in computer security.

    \item We conduct experiment and demonstrate the effectiveness of unsupervised approach using an ensemble method to compute anomaly scores for a large-scale enterprise dataset, enabling the identification of suspicious activities without extensive manual labeling.

    \item We evaluate the performances of supervised fine-tuned models on a few-shot set of labeled sessions, highlighting the adaptability and accuracy of our supervised approach.
\end{enumerate}

The remainder of this paper is organized as follows: Section 2 provides related work in command shell anomaly detection; Section 3 presents the data, including dataset description, differences from previous datasets, data quality issues, and data cleaning procedures; Section 4 details our methodology, including the unsupervised and supervised approaches; Section 5 presents the experimental results and examples of suspicious activities; and Section 6 concludes the paper and outlines possible future work.

\section{Related Work}

In this section, we discuss the existing literature related to detecting anomalies in Unix shell commands. We first review research in log anomaly detection and then masquerade detection. We also highlight the gaps in previous research that our proposed approach aims to address.

\subsection{Log Anomaly Detection}

Log anomaly detection \cite{yadav2020survey,le2022log} is an essential aspect of computer security since system logs provide important information about system activity and user behavior. By identifying anomalous patterns in logs, security analysts can detect potential threats, investigate incidents, and prevent service interruptions and data breaches. Deep learning techniques, including Long Short-Term Memory (LSTM) networks \cite{hochreiter1997long} and transformers \cite{vaswani2017attention}, have been applied to log anomaly detection, such as DeepLog \cite{du2017deeplog}, LogRobust \cite{zhang2019robust}, LogBERT \cite{guo2021logbert}, and NeuralLog \cite{le2021log}, demonstrating their ability to learn intricate patterns and long-range dependencies. 

However, these methods are primarily designed for analyzing system logs, which tend to be close to human languages in terms of syntax and semantics. In contrast, Unix shell commands exhibit distinct patterns and structures that may not be effectively captured by existing log anomaly detection approaches and pretrained language models. This limitation highlights the need for specialized methods tailored to command shell anomaly detection.

\subsection{Masquerade Detection}

Masquerade detection \cite{bertacchini2008survey} is a specific type of anomaly detection that focuses on identifying unauthorized users who have gained access to legitimate user's accounts or privileges and are attempting to impersonate them. The goal is to detect differences in user behavior between sessions that may indicate the presence of an attacker. In the context of Unix shell sessions, masquerade detection aims to distinguish between the normal activities of the genuine user and the suspicious actions of the masquerader. Early approaches to masquerade detection relied on traditional machine learning techniques, such as  Naive Bayes \cite{maxion2002masquerade,maxion2003masquerade,wang2003one}, Support Vector Machines (SVMs) \cite{wang2003one,kim2005empirical}, and Hidden Markov Models (HMM) \cite{liu2020hmms}. Deep learning techniques \cite{elmasry2018deep,yuan2021deep}, including Convolutional Neural Networks \cite{azeezat2021conceptual}, Temporal Convolutional Networks \cite{zhai2022masquerade}, and LSTM \cite{azeezat2021conceptual}, have also been applied to masquerade detection, leading to improved detection accuracies. 

However, these masquerade detection methods are not well-suited for detecting suspicious activities in Unix shell sessions. The goal of masquerade detection is to find imitators, while the command shell anomaly detection is trying to search suspicious or exploitable command patterns. Besides, the supervised method used in previous research can only detect anomalous sessions based on predefined rules and features from experts, which limit their flexibility and adaptability and make it challenging to identify new or unknown threats in command shell sessions.

\section{Data}

In this section, we describe the data used for our study, including the data description and data preprocessing. Important steps for extracting and cleaning commands from the raw keystroke data are highlighted. We also discuss the characteristics of the data that make it different from previous Unix shell datasets.

\subsection{Data Description}

Previous datasets for Unix shell commands include the SEA dataset \cite{schonlau2001computer}, Greenberg dataset \cite{greenberg1988using}, PU dataset \cite{lane1997application}, and NL2Bash \cite{lin2018nl2bash}. The SEA dataset, introduced by Schonlau et al. \cite{schonlau2001computer}, is a widely recognized benchmark, consisting of Unix commands from 50 users, with potential masquerade attacks seeded. The Greenberg dataset, collected by Greenberg et al. \cite{greenberg1988using}, contains Unix commands from 168 different users of the Unix C shell, and has been used to study user behavior and evaluate masquerade detection models. The PU dataset, developed by Lane et al. \cite{lane1997application}, contains 9 sets of sanitized user data collected from Purdue university command histories of 8 users in 2 years. The NL2Bash dataset, collected by Lin et al. \cite{lin2018nl2bash}, contains around 10,000 English sentence and bash command pairs. These datasets have contributed significantly to the development and evaluation of various Unix shell anomaly detection techniques, especially in the masquerade detection area. While each dataset offers unique insights, they also have their limitations, such as being outdated, only with truncated commands but without command options and subshells, lacking diversity of command usages, or not providing sufficient data for certain types of real exploits or attacks. Consequently, our study aims to leverage a large-scale, unlabeled dataset of Unix shell commands from real operating system users to explore novel anomaly detection approaches and address the limitations of previous datasets.

The raw data used in the research includes 90 days of Unix keystroke sessions from over 15,000 users, which have about 3 million activity objects. Among these activities, around 2.4 million objects are non-empty interactive sessions. However, the raw data have several data characteristics, including mixed shell prompts, command inputs, and command outputs, various shell prompts across sessions and within session, truncated long command lines with varying line lengths, various command aliases across sessions, mixed background process outputs with prompts and inputs, and missed backspaces and tab keys. In order to prepare this dataset for detecting anomalies in the next step, we developed heuristics to extract and clean commands from the raw data.

\subsection{Data Preprocessing}

The anomaly detector for shell commands needs clean command sessions to avoid introducing much noise into the model. However, the raw keystroke log dataset is a mixture of commands inputted by users and also responses outputted from systems. In order to increase the anomaly detection accuracies and also decrease the computing time, we extract user command inputs from the raw data and clean these commands. A heuristic algorithm is developed for this data preprocessing function, which is introduced briefly as follows.

In order to extract commands from the raw data, we need to search the shell prompts first. One conventional way is using the regular expressions. However, in practice, different sessions can have different shell prompts, and even in one session, the shell prompt can vary based on current working directories or subshells. Handcrafting regular expressions for each session is a tedious and non-adaptive work. To overcome these drawbacks, we create a list of 140 common Unix commands and a list of prompt terminal symbols (\$, \#, >). More terminal symbols were tested, but the probability of mismatching increased.  For each input line, the first occurring prompt terminal symbol is located, and the following word is tested against the common command set.  If this word is a known common command, the prompt is saved, otherwise it is skipped. To avoid mismatching prompts, several rules are applied for fixing corner cases, such as removing time prefixes, checking for balanced brackets in each prompt, and excluding environment variables.

After extracting session prompts, we then extract commands from the raw data, where we search for known prompts from this session and then extract the command line after the prompt. Additional steps are applied for handling several special cases, such as removing text editor buffers and concatenating wrapped multiple-line commands. Some meta data are also collected for down-stream use, including numbers of output lines and error messages. After extracting commands and dropping duplicates, we obtain 1.15 million sessions.

The last step is the command cleaning process. The main goal of this step is to reduce the data noise, so the anomaly detection model can give more precise results. We apply several filters for cleaning the extracted shell commands, including removing command lines with error messages, dropping command editing buffers and shell completions, deleting long consecutive spaces and over-repeated characters, filtering command names with regular expressions, masking numbers and special words, and cleaning cyclic commands usually generated by loops from shell scripts. The cleaned command shell sessions are then used in the next stage for both unsupervised an supervised approaches.

\section{Methodology}

In this section, we outline the methodology of our proposed anomaly detection approach for Unix shell sessions. Our approach employs both unsupervised and supervised learning techniques. We provide a detailed description of the unsupervised ensemble anomaly detector based on the pretrained DistilBERT model and also the supervised fine-tuning of the DistilBERT model using a few labeled data.

\subsection{Unsupervised Approach}

The unsupervised approach of our research involves pretraining a DistilBERT \cite{sanh2019distilbert} model from Hugging Face \cite{wolf2020transformers} on Unix shell commands and constructing an ensemble anomaly detector based on the session embeddings from the pretrained DistilBERT. This method was first proposed by CrowdStrike \cite{cocea2022bert,popa2022bert} for command lines from various platforms. The unsupervised model discovers new anomaly patterns for manual review.

Since the Unix shell commands are different from human languages, we pretrain a language model from scratch with the Unix shell commands instead of using an already existing pretrained model. BERT \cite{devlin2018bert} and its lighter-weight variant DistilBERT \cite{sanh2019distilbert} are state-of-the-art encoder-based transformer models that have shown remarkable performance in various natural language processing tasks, especially in understanding context and capturing complex language patterns. DistilBERT \cite{sanh2019distilbert} is selected in this research due to its balance of performance and efficiency. The WordPiece \cite{devlin2018bert}, the default sub-word tokenizer for DistilBERT, with a dictionary size of 30,000 is trained for tokenizing the Unix sessions, while several other dictionary sizes were experimented. Then the tokens are inputted into the DistilBERT model, and the model is pretrained for the masked language modeling (MLM) task to capture the inherent structure and dependencies within command sequences. The cased DistilBERT model is selected since the Unix shell is case-sensitive. This unsupervised pretraining allows the model to learn general representations of command sequences without relying on labeled data. Once the DistilBERT mode has been pretrained, the last hidden states are used as the embeddings of the Unix shell sessions. At the end of the pretraining process, we have one contextual embedding for each command session, which represents the higher-level features of the command sequences.

To detect anomalies of Unix sessions in an unsupervised approach without fine-tuning a classification layer, four outlier detectors from PyOD \cite{zhao2019pyod} are applied, including the principal component analysis (PCA) \cite{aggarwal2016outlier,shyu2003novel}, isolation forest (IF) \cite{liu2008isolation,liu2012isolation}, copula-based outlier detection (COPOD) \cite{li2020copod}, and autoencoders (AE) \cite{aggarwal2016outlier}, by following CrowdStrike’s framework \cite{cocea2022bert,popa2022bert}. These four outlier detection models are trained with the session embeddings, and their decision scores are normalized for each outlier detector. For each session, all four decision scores are averaged to get the final anomaly score of that session. The anomaly scores represent how deviant of one command session from the overall collection of sessions. Sessions with high anomaly scores are considered outliers, which may contain unusual command syntaxes or patterns. 

\subsection{Supervised Approach}

The supervised part of our approach involves fine-tuning the pretrained DistilBERT model with labeled data to improve its performance in distinguishing between normal and suspicious command sequences as a binary classifier. We fine-tune the pretrained DistilBERT with SetFit (Sentence Transformer Fine-tuning) \cite{tunstall2022efficient}, which is an efficient and prompt-free framework for few-shot fine-tuning of sentence transformers. In SetFit, the transformer can be fine-tuned on a small number of text pairs in a contrastive Siamese manner with high accuracy. The results of the model fine-tuned by SetFit are compared with the original fine-tuned DistilBERT and a trained logistic regressor with fixed session embedding. 

In order to fine-tune the pretrained model, examples of labeled sessions are required. Instead of labeling sessions manually, we create a table of suspicious keywords developed based on Uptycs’s work \cite{salunkhe2021linux} to cover MITRE ATT\&CK\textsuperscript{\textregistered} techniques \cite{mitre2023mitre,red20231atomic} commonly used by attackers. Those suspicious keywords are presented in the Table \ref{tab:keywords} with their corresponding technique IDs and names. Those suspicious keywords are searched in each Unix shell sessions, and those sessions with the number of unique suspicious keywords higher than the threshold are considered as anomalies. The setting of the labeled dataset is discussed further in the experimental results. Besides the suspicious keywords, we also created regular expressions to tag sessions with more ATT\&CK techniques \cite{mitre2023mitre,red20231atomic}. Those tags are used for the session annotation and analysis.

\begin{table*}[!h]
    \centering
    \footnotesize
    \caption{Suspicious keywords and MITRE ATT\&CK\textsuperscript{\textregistered} techniques.}
    \label{tab:keywords}
    \begin{tabular}{p{0.15\linewidth}p{0.4\linewidth}p{0.35\linewidth}}
        \toprule
        ATT\&CK Technique ID	&ATT\&CK Technique Name	&Suspicious Keywords \\
        \midrule
        T1018	&Remote System Discovery	&arp, ping, ip, hosts \\
        T1033	&System Owner/User Discovery	&whoami, who, w, users, USER \\
        T1049	&System Network Connections Discovery	&netstat, lsof, who, w \\
        T1016	&System Network Configuration Discovery	&arp, ipconfig, ifconfig, nbtstat, netstat, route, ping, ip \\
        T1082	&System Information Discovery	&df, uname, hostname, env, lspci, lscpu, lsmod, dmidecode, systeminfo \\
        T1087	&Account Discovery	&id, groups, lastlog, ldapsearch \\
        T1069	&Permission Groups Discovery	&groups, id, ldapsearch \\
        T1040	&Network Sniffing	&tcpdump, tshark \\
        T1574.006	&Hijack Execution Flow: Dynamic Linker Hijacking	&ld.so.preload, LD\_PRELOAD \\
        T1547.006	&Boot or Logon Autostart Execution: Kernel Modules and Extensions	&modprobe, insmod, lsmod, rmmod, modinfo \\
        T1136	&Create Account	&useradd, adduser \\
        T1053.003	&Scheduled Task/Job: Cron	&crontab, cron \\
        T1489	&Service Stop	&kill, pkill \\
        T1562.001	&Impair Defenses: Disable or Modify Tools	&systemctl \\
        T1105	&Ingress Tool Transfer	&curl, scp, sftp, tftp, rsync, finger, wget \\
        T1222.002	&File and Directory Permissions Modification: Linux and Mac File and Directory Permissions Modification	&chown, chmod, chgrp, chattr \\
        T1003.008	&OS Credential Dumping: /etc/passwd and /etc/shadow	&passwd, shadow \\
        T1070.003	&Indicator Removal: Clear Command History	&.bash\_history, HISTFILE, HISTFILESIZE \\
        T1548.003	&Abuse Elevation Control Mechanism: Sudo and Sudo Caching	&sudo, sudoers \\
        T1546.004	&Event Triggered Execution: Unix Shell Configuration Modification	&profile, profile.d, .profile, .bash\_profile, .bash\_login, .bashrc, .bash\_logout \\
        \bottomrule
    \end{tabular}
\end{table*}

Upon completing the supervised fine-tuning phase, we evaluate the performance of our anomaly detection approach using the testing data. We assess the model's effectiveness in detecting normal and suspicious command sequences by calculating various performance metrics, including precision, recall, and F1 score. The evaluations are discussed in the next section.

\section{Experimental Results}

In this section, we present the experimental results for both unsupervised and supervised anomaly detection methods applied to Unix shell commands. We first evaluate the unsupervised model with the pretrained DistilBERT embedding and the ensemble anomaly detector on the unlabeled data and then evaluate performance of the supervised model with labeled sessions.

\subsection{Unsupervised Approach Results}

In order to evaluate the unsupervised model and understand its performance, several analyses are done, including visualizing distributions of anomaly scores and embedding vectors, investigating relations between the anomaly scores and numbers of tokens and command lines, and also comparing anomaly scores of the common shell commands.

\begin{figure}[!h]
  \centering
  \begin{minipage}[b]{0.45\textwidth}
     \centering
     \includegraphics[width=\textwidth]{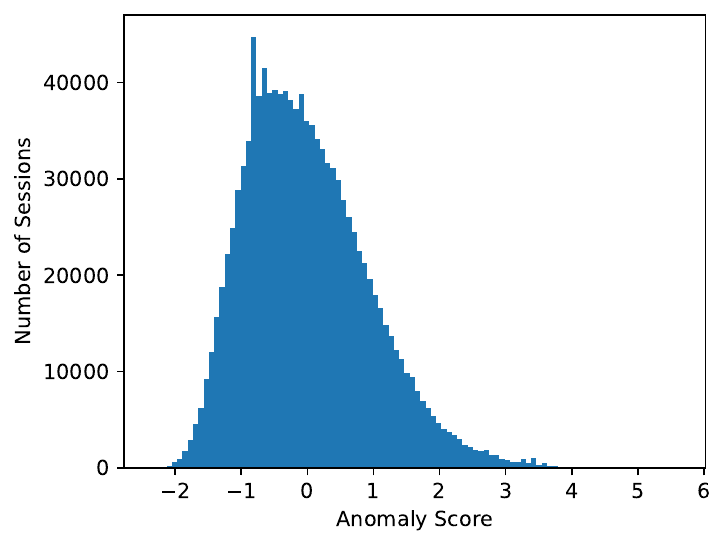}
     \caption{Distribution of averaged anomaly scores.}
    \label{fig:anomaly_scores}
  \end{minipage}
  \hfill
  \begin{minipage}[b]{0.5\textwidth}
     \centering
     \includegraphics[width=\textwidth]{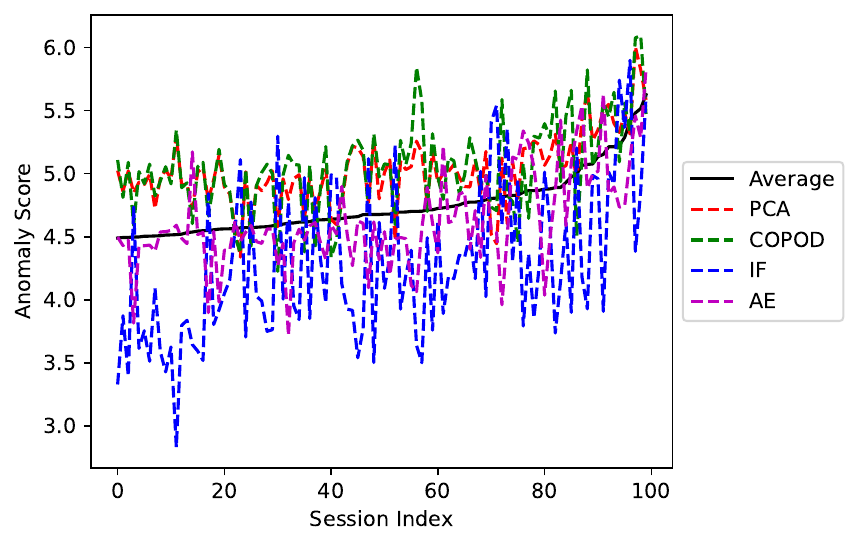}
     \caption{Four anomaly scores sorted by the averaged anomaly score.}
        \label{fig:all_scores}
  \end{minipage}
\end{figure}

The distribution of anomaly scores is shown in the Figure \ref{fig:anomaly_scores}. Since the anomaly scores have already been standardized, the mean and standard deviation of the distribution are 0 and 1 respectively. The distribution of anomaly scores is close to normal distribution, where most of sessions are observed around mean, while some outliers have higher anomaly scores than the most sessions. Besides, the anomaly scores from four anomaly detectors for the top 100 anomalies are also shown in the Figure \ref{fig:all_scores}, where the COPOD usually have the highest anomaly scores, while the IF tends to be the lower side and with a higher variance. For most sessions, these four anomaly detectors show consistent behaviors and assign high anomaly scores to these sessions.

\begin{figure}[!h]
  \centering
  \begin{minipage}[b]{0.45\textwidth}
    \centering
    \includegraphics[width=\textwidth]{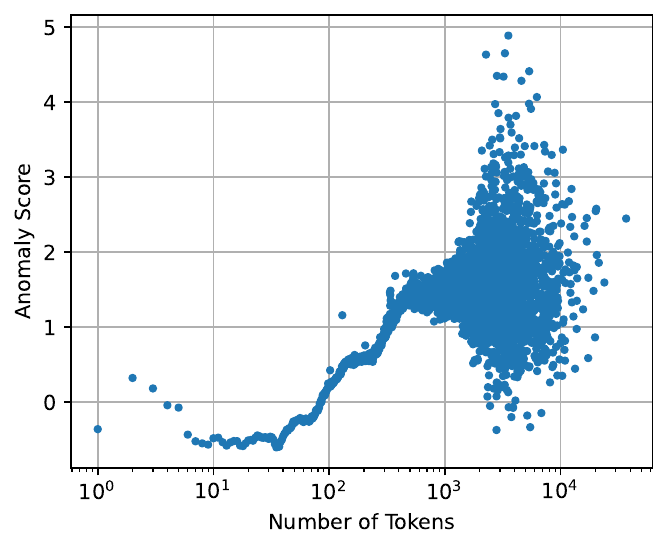}
    \caption{The relation between the number of tokens and the anomaly score.}
    \label{fig:num_tokens}
  \end{minipage}
  \hfill
  \begin{minipage}[b]{0.45\textwidth}
    \centering
    \includegraphics[width=\textwidth]{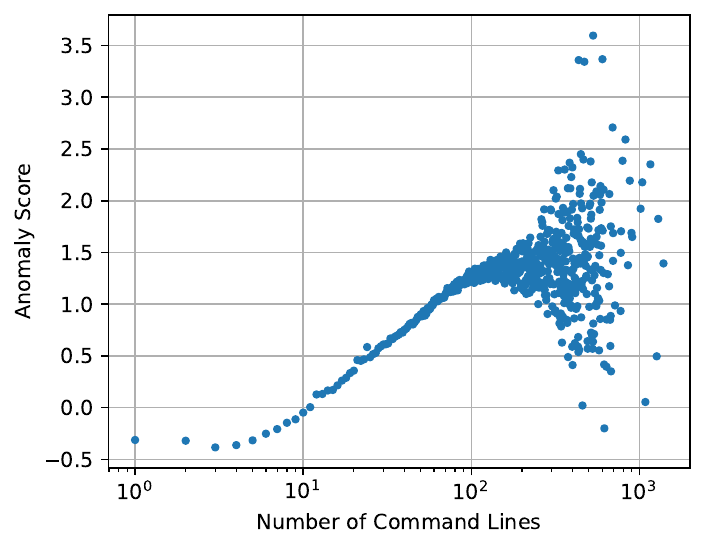}
    \caption{The relation between the number of command lines and the anomaly score.}
    \label{fig:num_lines}
  \end{minipage}
\end{figure}

To further understand the behavior of the unsupervised model, the anomaly scores are presented with the number of tokens and the number of command lines in the Figure \ref{fig:num_tokens} and Figure \ref{fig:num_lines}. Generally speaking, a session with more tokens and more command lines can have higher anomaly score. It is because usually shorter sessions only have the simple syntax for straightforward and repetitive daily usages, while longer sessions can have long command sequences to perform complicated and uncommon tasks, which are preferred by the unsupervised model due to their unusual command structure and syntax.

\begin{figure}[!h]
  \centering
  \includegraphics[width=\textwidth]{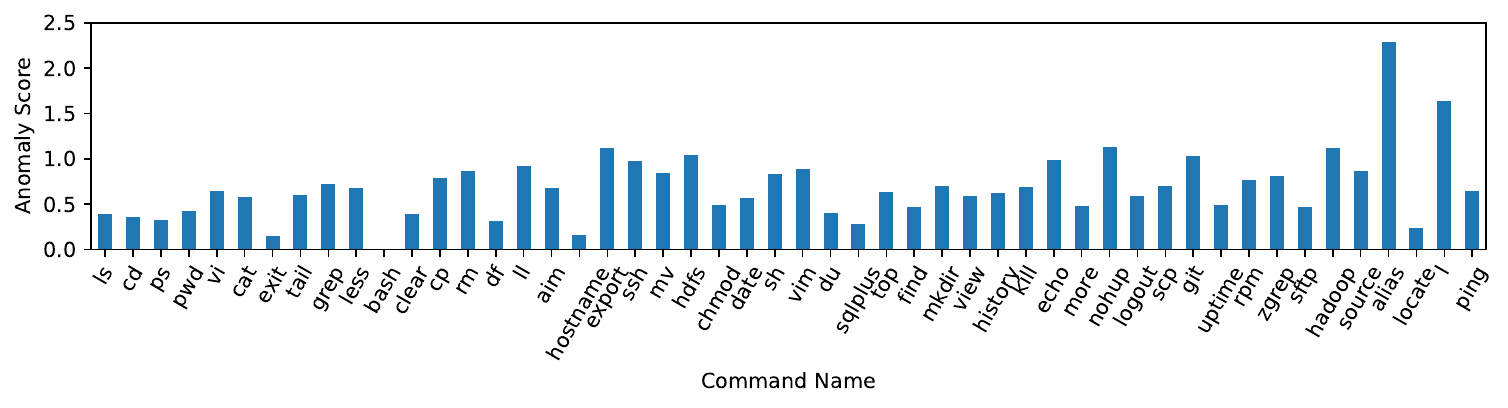}
  \caption{Averaged anomaly score for common command names.}
  \label{fig:command_names}
\end{figure}

At the end of unsupervised model analysis, we show the anomaly scores for the top 50 common commands in the Figure \ref{fig:command_names}. Those anomaly scores are weighted averaged of the session anomaly scores, where these commands appear. Most common commands, such as “ls” “exit”, “bash”, and so on, have lower anomaly scores, while “alias” and “l” have higher anomaly scores. In most cases, there is no clear explanation about the relation between the command names and their anomaly scores, since those anomaly scores are averaged from their sessions and can be affected by the session structures. But in general, infrequent commands have higher anomaly scores.

In summary, a session with a high anomaly score does not always mean it has the suspicious activity. However, anomaly scores can be used for prioritizing command sessions for expert analyses and also help monitoring experts discover new suspicious patterns. The unclear relations and uncertainties of the unsupervised model results motivate us to build and evaluate supervised models, which are discussed next. More investigation of relations between anomaly scores and suspicious activities and also the language structure of shell commands can be done in the future research.

In addition to the Unix shell, similar analyses are also done for subshell commands. During the command cleaning, we removed subshells which have different prompts than the Unix shells, such as HDFS, Spark, SQL, and Python. Those subshells are extracted separately, where an unsupervised anomaly detector in the same structure is applied to each subshell. The anomaly scores are assigned to subshell sessions, where specific exploits are also scanned through them. Analyzing the experiment results from subshell anomaly detection is beyond the scope of this paper.

\subsection{Supervised Approach Results}

To evaluate the supervised models, we label the command sessions by the number of suspicious keywords as described in the methodology. If one session has at least three unique suspicious keywords, it is considered as an abnormal session. However, if one session has zero suspicious keywords, it is labeled as a normal session. Other sessions are labeled as the abstained session, which are removed from model evaluations, since there is no strong criterion to classify them into either class. The labeled dataset is split into the training and testing sets by 90:10, and the number of sessions in each class are shown in the Table \ref{tab:classes}. During experiments, we use the same number of normal and abnormal sessions from the training data and combine them into a few-shot training set.

\begin{table*}[!h]
    \centering
    \footnotesize
    \caption{Number of sessions in the normal, abnormal, and abstained classes.}
    \label{tab:classes}
    \begin{tabular}{p{0.18\linewidth}p{0.25\linewidth}p{0.18\linewidth}p{0.12\linewidth}p{0.12\linewidth}}
        \toprule
        Class	&Number of Unique Suspicious Keywords	&Number of Samples	&Training Set (90\%)	&Testing Set (10\%) \\
        \midrule
        Normal	&= 0	&790,363	&711,327	&79,036 \\
        Abnormal	&>= 3	&28,413	&25,571	&2,842 \\
        Abstained (no label)	&In between	&335,322	&-	&- \\
        Total	&-	&1,154,098	&736,898	&81,878 \\
        \bottomrule
    \end{tabular}
\end{table*}

\begin{figure}[!h]
  \centering
  \includegraphics[width=0.5\textwidth]{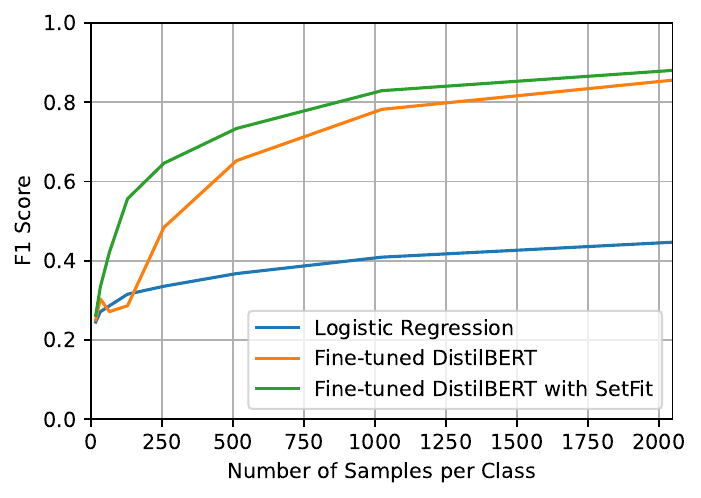}
  \caption{F1 scores of three supervised models with different training sizes.}
  \label{fig:f1_scores}
\end{figure}

Since the evaluation results from a small training set is unstable, we run 5 experiments for each model and each number of samples per class. For models fine-tuned with SetFit \cite{tunstall2022efficient}, we use the batch size 16, learning rate 1e-5, number of iterations 20 (number of text pairs), and train each model for 1 epoch. For fine-tuned DistilBERT models, we use the learning rate 1e-5, and each model is trained for 5 epochs. The averaged precisions, recalls, and F1 scores are reported in the Figure \ref{fig:f1_scores} and Table \ref{tab:results}. The fine-tuned SetFit model with 2048 samples per class shows the best result, which is higher than the fine-tuned DistilBERT with the same training data size. The fixed DistilBERT embedding with logistic regression gives the lowest result. The observation shows the advantage of SetFit for fine-tuning pretrained models when the labeled data are limited. Also, the model performance increases as the number of samples per class increasing. The experimental results of supervised model show the feasibility of creating a small set of manually labeled command sessions, fine-tuning a pretrained model with SetFit, and then using it for classifying more sessions automatically.

\begin{table*}[!h]
    \centering
    \footnotesize
    \caption{Evaluation results of three supervised models with different training sizes.}
    \label{tab:results}
    \begin{tabular}{c|ccc|ccc|ccc}
        \toprule
        Model & \multicolumn{3}{|c|}{Logistic Regression} & \multicolumn{3}{|c|}{Fine-tuned DistilBERT} & \multicolumn{3}{|c}{Fine-tuned DistilBERT with SetFit} \\
        \midrule
        Number of	&Precision	&Recall	&F1 Score	&Precision	&Recall	&F1 Score	&Precision	&Recall	&F1 Score \\
        Samples	&&&&&&&&& \\
        per Class	&&&&&&&&& \\
        \midrule
        16	&0.1464	&0.7860	&0.2454	&0.1632	&0.5578	&0.2513	&0.1569	&0.8287	&0.2622 \\
        32	&0.1625	&0.8248	&0.2711	&0.1995	&0.6977	&0.3036	&0.2059	&0.8930	&0.3331 \\
        64	&0.1713	&0.8754	&0.2862	&0.1625	&0.8418	&0.2716	&0.2712	&0.9484	&0.4210 \\
        128	&0.1922	&0.8849	&0.3155	&0.1703	&0.9098	&0.2864	&0.3909	&0.9758	&0.5563 \\
        256	&0.2070	&0.8890	&0.3356	&0.3230	&0.9663	&0.4840	&0.4819	&0.9850	&0.6459 \\
        512	&0.2308	&0.9027	&0.3676	&0.4900	&0.9774	&0.6524	&0.5845	&0.9866	&0.7337 \\
        1024	&0.2631	&0.9188	&0.4090	&0.6483	&0.9854	&0.7819	&0.7134	&\textbf{0.9900}	&0.8290 \\
        2048	&0.2944	&0.9267	&0.4467	&0.7534	&0.9899	&0.8555	&\textbf{0.7934}	&0.9894	&\textbf{0.8802} \\
        \bottomrule
    \end{tabular}
\end{table*}

\subsection{Session Annotations and Examples}

Besides experiments and evaluations of unsupervised and supervised models, we also annotated sessions with MITRE ATT\&CK\textsuperscript{\textregistered} techniques in addition to previously mentioned suspicious keywords and anomaly scores. These annotations can help cybersecurity experts recognize and analyze suspicious activity. 

\begin{figure}[!h]
  \centering
  \includegraphics[width=\textwidth]{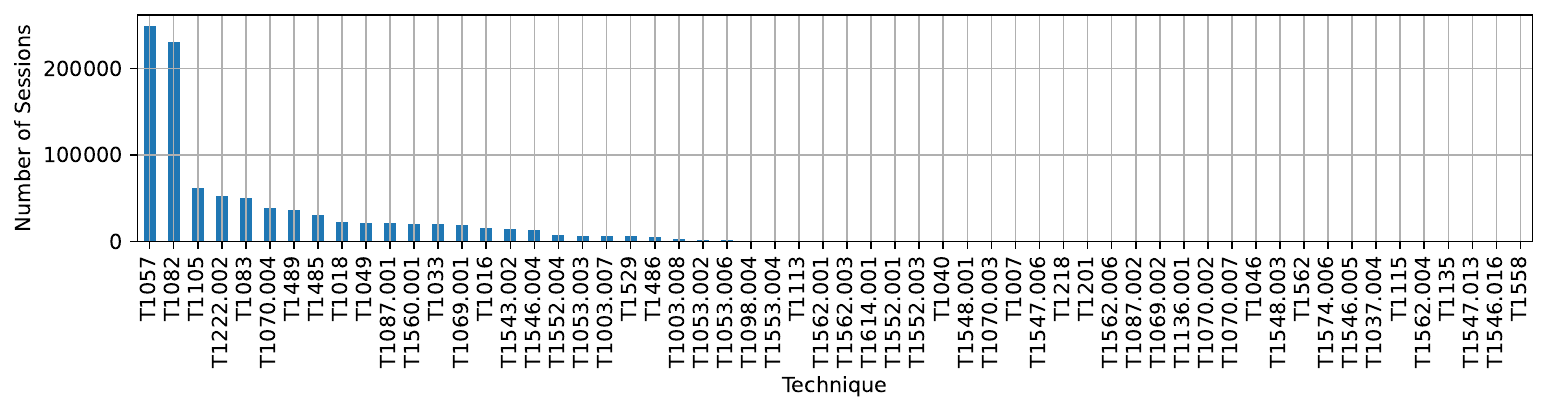}
  \caption{Number of sessions for different MITRE ATT\&CK\textsuperscript{\textregistered} techniques.}
  \label{fig:techniques}
\end{figure}

\begin{figure}[!h]
  \centering
  \includegraphics[width=0.6\textwidth]{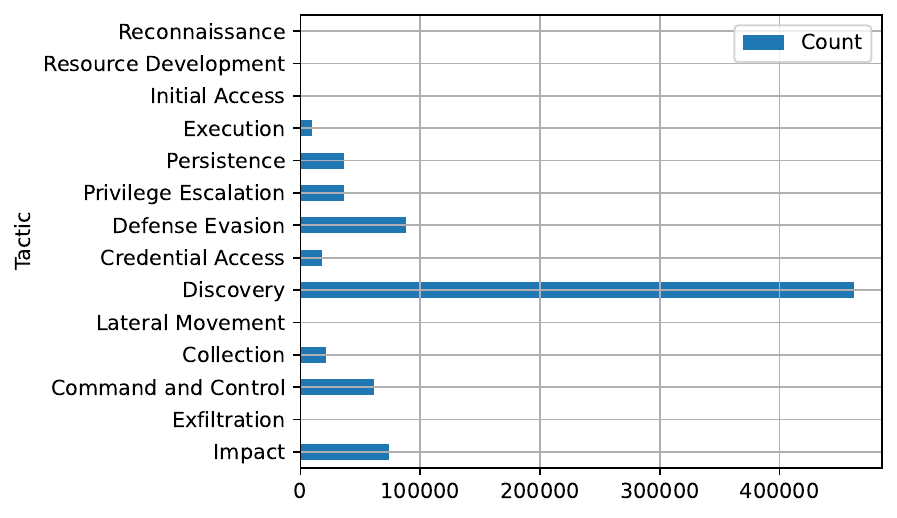}
  \caption{Number of sessions for different MITRE ATT\&CK\textsuperscript{\textregistered} tactics.}
  \label{fig:tactics}
\end{figure}

During the annotation process, Unix shell sessions are labeled by searching 58 MITRE ATT\&CK\textsuperscript{\textregistered} techniques with corresponding regular expressions. For each technique, we search for specific command usages and file accesses. The distributions of techniques are shown in Figure \ref{fig:techniques}, and the tactics are shown in Figure \ref{fig:tactics}. The most common techniques are T1057 Process Discovery, T1082 System Information Discovery, and T1105 Ingress Tool Transfer, although those sessions with less-common techniques are more interesting to be analyzed for anomaly detection. 

Three session examples with high anomaly scores are selected and presented in the Table \ref{tab:wget}-\ref{tab:chmod}, where ATT\&CK techniques are highlighted in the blue color with suspicious keywords in the red color. The first example in the Table \ref{tab:wget} shows remote command execution of transient web server with potential for data exfiltration. The second example in the Table \ref{tab:credential} gives a potential data exfiltration and credential exposure subject to discovery via process discovery. And the last example in the Table \ref{tab:chmod} illustrates disk clear and boot load configuration changes.

\begin{table*}[!h]
    \caption{An example of remote command execution of transient web server with potential for data exfiltration.}
    \label{tab:wget}
    \centering
    \footnotesize
    \begin{tabular}{p{0.02\linewidth}|p{0.85\linewidth}}
        \toprule
        \multicolumn{2}{l}{Activity id = *1e1BD9. \textbf{Anomaly score = 1.8919.} Suspicious keywords = [kill: 3, wget: 21]} \\
        \midrule
        1 & \texttt{<lines removed>} \\
        2 & \texttt{salt "WH" cmd.run "python -m SimpleHTTPServer \# --directory /sqldata/ms\_backups/" bg=trues/WH\_test\_db\_FU} \\
        3 & \texttt{salt "WH" cmd.run "ps aux | grep '[S]impleHTTPServer \#' | awk '\{print \$\#\}' |xargs {\color{red}kill} –9 "/WH\_test\_db\_FUWH:} \\
        4 & \texttt{\color{blue}-> [T1057: Process Discovery, T1489: Service Stop]} \\
        5 & \texttt{salt "WH" cmd.run "cd /sqldata/dbmigration;{\color{red}wget} http://<host:port>//sqldata/ms\_backups/WH\_test\_db\_FU} \\
        6 & \texttt{\color{blue}-> [T1105: Ingress Tool Transfer]} \\
        7 & \texttt{<lines removed>} \\
        \midrule
        \multicolumn{2}{l}{Details:} \\
        \multicolumn{2}{l}{Line 2: launch transient web server on remote host.} \\
        \multicolumn{2}{l}{Line 3: terminate the server.} \\
        \multicolumn{2}{l}{Line 4 and 6: ATT\&CK tags inserted by processing pipeline.} \\
        \multicolumn{2}{l}{Line 5: transfer data from web server using wget.} \\
        \bottomrule
    \end{tabular}
\end{table*}

\begin{table*}[!h]
    \caption{An example of potential data exfiltration and credential exposure subject to discovery via process discovery.}
    \label{tab:credential}
    \centering
    \footnotesize
    \begin{tabular}{p{0.02\linewidth}|p{0.85\linewidth}}
        \toprule
        \multicolumn{2}{l}{Activity id = *1c01C8. \textbf{Anomaly score = 1.9754.} Suspicious keywords = [curl: 12]} \\
        \midrule
        1 & \texttt{<lines removed>} \\
        2 & \texttt{{\color{red}curl} -T server\_support.tar.gz -u<username>:<plaintext\_credentials> <externalhost> /dropzone/uploads} \\
        3 & \texttt{\color{blue}-> [T1105: Ingress Tool Transfer]} \\
        4 & \texttt{<lines removed>} \\
        \bottomrule
    \end{tabular}
\end{table*}

\begin{table*}[!h]
    \caption{An example of disk clear and boot load configuration changes.}
    \label{tab:chmod}
    \centering
    \footnotesize
    \begin{tabular}{p{0.02\linewidth}|p{0.85\linewidth}}
        \toprule
        \multicolumn{2}{l}{Activity id = *b41A0E. \textbf{Anomaly score = 3.1271.} Suspicious keywords = [chmod: 2, df: 1, wget: 1]} \\
        \midrule
        1 & \texttt{< lines removed >}\\
        2 & \texttt{ansible all -i <INVENTORY> -m shell -a "uptime;grep Start /etc/INSTALL\_CLASS;cat /etc/redhat-release" -o>}\\
        3 & \texttt{\color{blue}-> [T1082: System Information Discovery]>}\\
        4 & \texttt{ansible all -i <INVENTORY> -m shell -a "cd /root;{\color{red}chmod} HFF diskwipe.sh;./diskwipe.sh" -b>}\\
        5 & \texttt{\color{blue}-> [T1222.002: File and Directory Permissions Modification - Linux and Mac File and Directory Permissions Mod]>}\\
        6 & \texttt{ansible all -i <INVENTORY> -m shell -a "/sbin/service ambari-agent restart"  -become -b>}\\
        7 & \texttt{<lines removed>>}\\
        8 & \texttt{ansible all -i <INVENTORY> -m shell -a "cd /boot/grub\#;cp -p grub.cfg grub.cfg.bkp"  -b>}\\
        9 & \texttt{ansible all -i <INVENTORY> -m shell -a "/sbin/grubby --args=transparent\_hugepage=never --update-kernel=ALL "  -b>}\\
        10 & \texttt{<lines removed>}\\
        \midrule
        \multicolumn{2}{l}{Details:} \\
        \multicolumn{2}{l}{Lines 1, 7, 10 omitted for brevity.} \\
        \multicolumn{2}{l}{Line 3 and 5 are automatic annotations added by pipeline.} \\
        \multicolumn{2}{l}{Line 2: remote command to check system details.} \\
        \multicolumn{2}{l}{Line 4: remote command to clear disk prior to install.} \\
        \multicolumn{2}{l}{Line 6: restart Hadoop monitoring agent.} \\
        \multicolumn{2}{l}{Line 8, 9: modify boot loader.} \\
        \bottomrule
    \end{tabular}
\end{table*}

\section{Conclusions}

Anomaly detection for interactive command shells is a complex problem. Detection of anomalies is needed as a cybersecurity safeguard because privileged access at the shell level provides the opportunity for a range of attacks that threaten critical enterprise infrastructure, data, and services. On the other hand, prevention of such threats by locking system access prevents important operations activities like upgrades, change management, and outage investigation and remediation.

Prior research has been limited by available datasets. We presented the first published results on keystroke anomaly detection using an enterprise-scale dataset captured from production systems over a 90-day period. The extent of the dataset, 1.15 million sessions captured from over 15,000 users, demonstrates the need for automated anomaly detection. The dataset came with important data extraction and cleaning issues but provides a rich cross-section of enterprise operations activities.  Notably, the monitored infrastructure in the dataset excludes network appliances and specialized embedded systems and is otherwise representative of widely used information technology.

Past research has also been limited by available models. We presented the first experimental results of using a machine-learning transformer model, specifically DistilBERT, for keystroke log anomaly detection of Unix shells, in both unsupervised and supervised approaches. Although the dataset is unlabeled, we tagged each session using two existing schemes: the MITRE ATT\&CK\textsuperscript{\textregistered} techniques and suspicious keywords.  Unix shell sessions with high anomaly scores were then cross-checked with the tags as part of validating the utility of the anomaly model for operations uses. Model output was also compared with rule-based log analysis scripts used by operations teams. The results of the cross-check show that the outliers found by the model contain significant cases not found in either the tagging or existing analysis scripts. More future research can be done for designing specific tokenizers for shell commands, understanding the implicit relations between anomaly scores and suspicious activities, and analyzing subshell command anomalies.



\small

\begin{thebibliography}{10}
	
	\bibitem{aggarwal2016outlier}
	Charu~C. Aggarwal.
	\newblock {\em Outlier Analysis}.
	\newblock Springer Publishing Company, Incorporated, 2nd edition, 2016.
	
	\bibitem{azeezat2021conceptual}
	Adam~Adenike Azeezat, Onashoga~Sadiat Adebukola, Abayomi-Alli Adebayo, and
	Omoyiola~Bayo Olushola.
	\newblock A conceptual hybrid model of deep convolutional neural network (dcnn)
	and long short-term memory (lstm) for masquerade attack detection.
	\newblock In {\em Information and Communication Technology and Applications:
		Third International Conference, ICTA 2020, Minna, Nigeria, November 24--27,
		2020, Revised Selected Papers 3}, pages 170--184. Springer, 2021.
	
	\bibitem{bertacchini2008survey}
	Maximiliano Bertacchini and Pablo Fierens.
	\newblock A survey on masquerader detection approaches.
	\newblock In {\em Proceedings of V Congreso Iberoamericano de Seguridad
		Inform{\'a}tica, Universidad de la Rep{\'u}blica de Uruguay}, pages 46--60,
	2008.
	
	\bibitem{red20231atomic}
	Red Canary\textsuperscript{\textregistered}.
	\newblock Atomic red team\textsuperscript{\texttrademark}.
	\newblock \url{https://github.com/redcanaryco/atomic-red-team}, May 2023.
	\newblock Accessed: 2023-03-01.
	
	\bibitem{cocea2022bert}
	Stefan-Bogdan Cocea.
	\newblock Bert embeddings: A modern machine-learning approach for detecting
	malware from command lines (part 1 of 2).
	\newblock
	\url{https://www.crowdstrike.com/blog/bert-embeddings-new-approach-for-command-line-anomaly-detection/},
	January 2022.
	\newblock Accessed: 2022-06-01.
	
	\bibitem{mitre2023mitre}
	The~MITRE Corporation.
	\newblock Mitre att\&ck\textsuperscript{\textregistered} enterprise
	techniques,.
	\newblock \url{https://attack.mitre.org/techniques/enterprise}, 2023.
	\newblock Accessed: 2023-03-01.
	
	\bibitem{devlin2018bert}
	Jacob Devlin, Ming-Wei Chang, Kenton Lee, and Kristina Toutanova.
	\newblock Bert: Pre-training of deep bidirectional transformers for language
	understanding.
	\newblock {\em arXiv preprint arXiv:1810.04805}, 2018.
	
	\bibitem{du2017deeplog}
	Min Du, Feifei Li, Guineng Zheng, and Vivek Srikumar.
	\newblock Deeplog: Anomaly detection and diagnosis from system logs through
	deep learning.
	\newblock In {\em Proceedings of the 2017 ACM SIGSAC conference on computer and
		communications security}, pages 1285--1298, 2017.
	
	\bibitem{elmasry2018deep}
	Wisam Elmasry, Akhan Akbulut, and Abdul~Halim Zaim.
	\newblock Deep learning approaches for predictive masquerade detection.
	\newblock {\em Security and Communication Networks}, 2018, 2018.
	
	\bibitem{greenberg1988using}
	Saul Greenberg.
	\newblock Using unix: Collected traces of 168 users.
	\newblock Technical report, Research Report 88/333/45, Department of Computer
	Science, University of Calgary, Calgary, Alberta, 1988.
	
	\bibitem{guo2021logbert}
	Haixuan Guo, Shuhan Yuan, and Xintao Wu.
	\newblock Logbert: Log anomaly detection via bert.
	\newblock In {\em 2021 international joint conference on neural networks
		(IJCNN)}, pages 1--8. IEEE, 2021.
	
	\bibitem{hochreiter1997long}
	Sepp Hochreiter and J{\"u}rgen Schmidhuber.
	\newblock Long short-term memory.
	\newblock {\em Neural computation}, 9(8):1735--1780, 1997.
	
	\bibitem{kim2005empirical}
	Han-Sung Kim and Sung-Deok Cha.
	\newblock Empirical evaluation of svm-based masquerade detection using unix
	commands.
	\newblock {\em Computers \& Security}, 24(2):160--168, 2005.
	
	\bibitem{lane1997application}
	Terran Lane and Carla~E Brodley.
	\newblock An application of machine learning to anomaly detection.
	\newblock In {\em Proceedings of the 20th national information systems security
		conference}, volume 377, pages 366--380. Baltimore, USA, 1997.
	
	\bibitem{le2021log}
	Van-Hoang Le and Hongyu Zhang.
	\newblock Log-based anomaly detection without log parsing.
	\newblock In {\em 2021 36th IEEE/ACM International Conference on Automated
		Software Engineering (ASE)}, pages 492--504. IEEE, 2021.
	
	\bibitem{le2022log}
	Van-Hoang Le and Hongyu Zhang.
	\newblock Log-based anomaly detection with deep learning: How far are we?
	\newblock In {\em Proceedings of the 44th international conference on software
		engineering}, pages 1356--1367, 2022.
	
	\bibitem{li2020copod}
	Zheng Li, Yue Zhao, Nicola Botta, Cezar Ionescu, and Xiyang Hu.
	\newblock Copod: copula-based outlier detection.
	\newblock In {\em 2020 IEEE international conference on data mining (ICDM)},
	pages 1118--1123. IEEE, 2020.
	
	\bibitem{lin2018nl2bash}
	Xi~Victoria Lin, Chenglong Wang, Luke Zettlemoyer, and Michael~D Ernst.
	\newblock Nl2bash: A corpus and semantic parser for natural language interface
	to the linux operating system.
	\newblock In {\em Proceedings of the Eleventh International Conference on
		Language Resources and Evaluation (LREC 2018)}, 2018.
	
	\bibitem{liu2008isolation}
	Fei~Tony Liu, Kai~Ming Ting, and Zhi-Hua Zhou.
	\newblock Isolation forest.
	\newblock In {\em 2008 eighth ieee international conference on data mining},
	pages 413--422. IEEE, 2008.
	
	\bibitem{liu2012isolation}
	Fei~Tony Liu, Kai~Ming Ting, and Zhi-Hua Zhou.
	\newblock Isolation-based anomaly detection.
	\newblock {\em ACM Transactions on Knowledge Discovery from Data (TKDD)},
	6(1):1--39, 2012.
	
	\bibitem{liu2020hmms}
	Jia Liu, Miyi Duan, Wenfa Li, and Xinguang Tian.
	\newblock Hmms based masquerade detection for network security on with parallel
	computing.
	\newblock {\em Computer Communications}, 156:168--173, 2020.
	
	\bibitem{maxion2003masquerade}
	Roy~A Maxion.
	\newblock Masquerade detection using enriched command lines.
	\newblock In {\em 2003 International Conference on Dependable Systems and
		Networks, 2003. Proceedings.}, pages 5--5. IEEE Computer Society, 2003.
	
	\bibitem{maxion2002masquerade}
	Roy~A Maxion and Tahlia~N Townsend.
	\newblock Masquerade detection using truncated command lines.
	\newblock In {\em Proceedings international conference on dependable systems
		and networks}, pages 219--228. IEEE, 2002.
	
	\bibitem{popa2022bert}
	Cristian Popa.
	\newblock Bert embeddings: A modern machine-learning approach for detecting
	malware from command lines (part 2 of 2).
	\newblock
	\url{https://www.crowdstrike.com/blog/bert-embeddings-new-approach-for-command-line-anomaly-detection-part-2/},
	April 2022.
	\newblock Accessed: 2022-06-01.
	
	\bibitem{radford2018improving}
	Alec Radford, Karthik Narasimhan, Tim Salimans, Ilya Sutskever, et~al.
	\newblock Improving language understanding by generative pre-training.
	\newblock 2018.
	
	\bibitem{salunkhe2021linux}
	Pritam Salunkhe.
	\newblock Linux commands \& utilities commonly used by attackers.
	\newblock
	\url{https://www.uptycs.com/blog/linux-commands-and-utilities-commonly-used-by-attackers},
	May 2021.
	\newblock Accessed: 2022-10-01.
	
	\bibitem{sanh2019distilbert}
	Victor Sanh, Lysandre Debut, Julien Chaumond, and Thomas Wolf.
	\newblock Distilbert, a distilled version of bert: smaller, faster, cheaper and
	lighter.
	\newblock {\em arXiv preprint arXiv:1910.01108}, 2019.
	
	\bibitem{schonlau2001computer}
	Matthias Schonlau, William DuMouchel, Wen-Hua Ju, Alan~F Karr, Martin Theus,
	and Yehuda Vardi.
	\newblock Computer intrusion: Detecting masquerades.
	\newblock {\em Statistical science}, pages 58--74, 2001.
	
	\bibitem{shyu2003novel}
	Mei-Ling Shyu, Shu-Ching Chen, Kanoksri Sarinnapakorn, and LiWu Chang.
	\newblock A novel anomaly detection scheme based on principal component
	classifier.
	\newblock In {\em Proceedings of the IEEE foundations and new directions of
		data mining workshop}, pages 172--179. IEEE Press, 2003.
	
	\bibitem{tunstall2022efficient}
	Lewis Tunstall, Nils Reimers, Unso Eun~Seo Jo, Luke Bates, Daniel Korat, Moshe
	Wasserblat, and Oren Pereg.
	\newblock Efficient few-shot learning without prompts.
	\newblock {\em arXiv preprint arXiv:2209.11055}, 2022.
	
	\bibitem{vaswani2017attention}
	Ashish Vaswani, Noam Shazeer, Niki Parmar, Jakob Uszkoreit, Llion Jones,
	Aidan~N Gomez, {\L}ukasz Kaiser, and Illia Polosukhin.
	\newblock Attention is all you need.
	\newblock {\em Advances in neural information processing systems}, 30, 2017.
	
	\bibitem{wang2003one}
	Ke~Wang and Salvatore~J Stolfo.
	\newblock One-class training for masquerade detection.
	\newblock In {\em Workshop on Data Mining for Computer Security}, 2003.
	
	\bibitem{wolf2020transformers}
	Thomas Wolf, Lysandre Debut, Victor Sanh, Julien Chaumond, Clement Delangue,
	Anthony Moi, Pierric Cistac, Tim Rault, R{\'e}mi Louf, Morgan Funtowicz,
	et~al.
	\newblock Transformers: State-of-the-art natural language processing.
	\newblock In {\em Proceedings of the 2020 conference on empirical methods in
		natural language processing: system demonstrations}, pages 38--45, 2020.
	
	\bibitem{yadav2020survey}
	Rakesh~Bahadur Yadav, P~Santosh Kumar, and Sunita~Vikrant Dhavale.
	\newblock A survey on log anomaly detection using deep learning.
	\newblock In {\em 2020 8th International Conference on Reliability, Infocom
		Technologies and Optimization (Trends and Future Directions)(ICRITO)}, pages
	1215--1220. IEEE, 2020.
	
	\bibitem{yuan2021deep}
	Shuhan Yuan and Xintao Wu.
	\newblock Deep learning for insider threat detection: Review, challenges and
	opportunities.
	\newblock {\em Computers \& Security}, 104:102221, 2021.
	
	\bibitem{zhai2022masquerade}
	Haibin Zhai, Yong Wang, Xueqiang Zou, Yihan Wu, Songyue Chen, Hongwei Wu, and
	Yanqin Zheng.
	\newblock Masquerade detection based on temporal convolutional network.
	\newblock In {\em 2022 IEEE 25th International Conference on Computer Supported
		Cooperative Work in Design (CSCWD)}, pages 305--310. IEEE, 2022.
	
	\bibitem{zhang2019robust}
	Xu~Zhang, Yong Xu, Qingwei Lin, Bo~Qiao, Hongyu Zhang, Yingnong Dang, Chunyu
	Xie, Xinsheng Yang, Qian Cheng, Ze~Li, et~al.
	\newblock Robust log-based anomaly detection on unstable log data.
	\newblock In {\em Proceedings of the 2019 27th ACM Joint Meeting on European
		Software Engineering Conference and Symposium on the Foundations of Software
		Engineering}, pages 807--817, 2019.
	
	\bibitem{zhao2019pyod}
	Yue Zhao, Zain Nasrullah, and Zheng Li.
	\newblock Pyod: A python toolbox for scalable outlier detection.
	\newblock {\em arXiv preprint arXiv:1901.01588}, 2019.
	
\end{thebibliography}

\end{document}